\documentclass[conference]{IEEEtran}
\IEEEoverridecommandlockouts
\usepackage{cite}
\usepackage{amsmath,amssymb,amsfonts}
\usepackage{bbm}
\usepackage{graphicx}
\usepackage{textcomp}
\usepackage{xcolor}
\usepackage{lineno}
\usepackage{booktabs}
\usepackage{float}
\usepackage{placeins}
\usepackage{algorithm}
\usepackage{algpseudocode}
\usepackage{enumitem}
\usepackage{url}
\usepackage{tikz}
\usetikzlibrary{arrows.meta, positioning, shapes.geometric}

\def\BibTeX{{\rm B\kern-.05em{\sc i\kern-.025em b}\kern-.08em
    T\kern-.1667em\lower.7ex\hbox{E}\kern-.125emX}}
\begin{document}

\title{LLM Based Bayesian Optimization for Prompt Search}

\author{
\IEEEauthorblockN{Adam Ballew \qquad Jingbo Wang \qquad Shaogang Ren}
\IEEEauthorblockA{Department of Computer Science and Engineering\\
University of Tennessee at Chattanooga\\
615 McCallie Ave, Chattanooga, TN 37403 \\
Email: adam.ballew94@gmail.com \quad ydk297@mocs.utc.edu \quad shaogang-ren@utc.edu}
}

\maketitle

\begin{abstract}
Bayesian Optimization (BO) has been widely used to efficiently optimize expensive black-box functions with limited evaluations. In this paper, we investigate the use of BO for prompt engineering to enhance text classification  with Large Language Models (LLMs). We employ an LLM-powered Gaussian Process (GP) as the surrogate model to estimate the performance of different prompt candidates. These candidates are generated by an LLM through the expansion of a set of seed prompts and are subsequently evaluated using an Upper Confidence Bound (UCB) acquisition function in conjunction with the GP posterior. The optimization process iteratively refines the prompts based on a subset of the data, aiming to improve classification accuracy while reducing the number of API calls by leveraging the prediction uncertainty of the LLM-based GP. The proposed BO-LLM\footnote{\url{https://github.com/ShaogangRen/BO-LLM}}
 algorithm is evaluated on two datasets, and its advantages are discussed in detail in this paper.
\end{abstract}


\section{Introduction}

The advancements in Large Language Models (LLMs), such as OpenAI's GPT, have allowed their application in various Natural Language Processing (NLP) tasks, including text classification~\cite{wang2023largelanguagemodelszeroshot}, question answering~\cite{brown2020language}, and text generation~\cite{ouyang2022training}. 
However, the performance of these models is highly sensitive to the structure and wording of the prompt it is given, making prompt optimization an important problem to solve~\cite{KNOTH2024100225}.

To address this, we propose using Bayesian Optimization (BO), a powerful framework for derivative-free optimization of expensive black-box functions~\cite{brochu2010tutorialbayesianoptimizationexpensive}, to systematically refine prompts for binary classification tasks. BO has been successfully applied to hyperparameter tuning~\cite{snoek2012practical}, neural architecture search~\cite{kandasamy2018neural}, and model calibration~\cite{astudillo2021bayesian,ren2024causal}, demonstrating its effectiveness in high-dimensional discrete optimization problems. The key challenge lies in optimizing the prompt structure to maximize classification accuracy while minimizing the number of expensive API calls. BO offers an efficient strategy by modeling the relationship between different prompts and their classification performance using a Gaussian Process (GP) as a surrogate function~\cite{rasmussen2006gaussian}. 
The optimization process is guided by an Upper Confidence Bound (UCB) 
acquisition function~\cite{srinivas2010gaussian}, which balances exploration (testing new prompts with high uncertainty) and exploitation (selecting prompts with a high estimated accuracy) to iteratively select prompt candidates. 
Our BO-LLM framework combines this GP surrogate and UCB acquisition with an LLM-based expansion mechanism to generate candidates.

Our contributions are: (i) introducing BO-LLM as a black-box prompt 
optimization framework, (ii) evaluating it on LIAR~\cite{wang2017liarliarpantsfire} and ETHOS~\cite{Mollas_2022} against ProTeGi~\cite{pryzant2023automaticpromptoptimizationgradient}, 
and (iii) showing adaptability to multi-turn QA.

\section{Background}

\subsection{Prompt Engineering}
The quality of a prompt is crucial to LLM performance in fulfilling a user's request, with key aspects being its clarity, relevancy, and structure. The specific wording has been shown to greatly affect the quality of the response, and a prompt that clearly states the user's request with minimal irrelevant information generally yields better outcomes~\cite{amatriain2024promptdesignengineeringintroduction, KNOTH2024100225}.

Prompt engineering is the practice of refining a prompt until the LLM's output is as close to the desired outcome as possible. Historically, this has been a time-consuming process requiring human-driven trial-and-error, where the user's domain knowledge significantly determines the quality of the final prompt. An automated prompt engineering process can help to alleviate these problems by reducing or completely removing this manual loop and leveraging the LLM's vast knowledge to generate effective prompts.

LLMs interact with users through a prompt–response paradigm, where the prompt's formulation strongly influences the output's usefulness. For example, prompts that clearly state the request first before providing relevant context are consistently more effective. Given that manual prompt crafting is both time-consuming and dependent on user expertise, automated methods are desirable to improve efficiency and reproducibility. This motivates the use of systematic frameworks, such as BO, to automate the exploration of the prompt space.

\subsection{Bayesian Optimization}

BO is a sequential model-based optimization framework well-suited for optimizing black-box functions that are expensive to evaluate. As outlined by Brochu et al.~\cite{brochu2010tutorialbayesianoptimizationexpensive}, instead of directly optimizing the objective function, BO constructs a surrogate model—commonly a GP~\cite{rasmussen2006gaussian}—that approximates the objective based on prior observations. This probabilistic model provides not only an estimation of the function's value but also its uncertainty across the input space. The core steps are as follows:

\textbf{Step 1: Initialization.} A surrogate model, such as a GP, is chosen. The GP is defined by a mean function $\mu(x)$ and a kernel (or covariance function) $k(x,x')$, which computes the relationship between observation data point.
\begin{equation}
    f(x)\sim \mathcal{GP}(\mu(x),k(x,x'))
\end{equation}
An initial observation dataset $D=\{(x_i,y_i)\}_{i=1}^n$ is collected, where $x$ represents an input (a feature vector or a prompt) and $y$ is its corresponding evaluation from the objective function (accuracy).

\textbf{Step 2: Posterior Inference.} Given the current dataset $D$ (the prior), the GP is fitted to obtain a posterior distribution over the objective function. This posterior captures our updated belief about the function after observing the data.

\textbf{Step 3: Acquisition Function.}
We adopt the Upper Confidence Bound acquisition, which balances exploration and exploitation.
See Eq.~(\ref{eq:ucb}) in Sec.~III-F for the exact form used in our framework.
UCB also enjoys theoretical guarantees in the bandit setting~\cite{srinivas2010gaussian}.

\textbf{Step 4: Sample Point Selection.}
Choose the next point by maximizing the acquisition:
\[
x_{\text{next}} = \arg\max_{x} \alpha(x),
\]
where $\alpha(\cdot)$ is the UCB acquisition defined in Eq.~(\ref{eq:ucb}).

\textbf{Step 5: Objective Evaluation and Update.} The true objective function is evaluated at $x_{next}$ to get a new observation $y_{next}$. This new data point $(x_{next}, y_{next})$ is added to the dataset $D$, and the process repeats from Step 2\cite{brochu2010tutorialbayesianoptimizationexpensive}. This loop continues until a termination condition, such as a fixed number of iterations, is met.

\subsection{Related Work}
Beyond manual crafting, several automated prompt optimization techniques have emerged. 
One prominent approach is Automatic Prompt Engineer (APE)~\cite{zhou2023large}, which 
leverages an LLM to generate a diverse set of instruction candidates and then selects the best one based on a chosen scoring function. This paradigm has been further explored 
in Optimization by PROmpting (OPRO)~\cite{yang2023large}, where the LLM iteratively generates improved prompts based on past performance feedback. Unlike our model-based approach, these methods rely on a generate-and-test loop without an explicit surrogate model capturing prompt-performance relationships.

Another line of work frames prompt optimization as a discrete search problem. 
RLPrompt~\cite{deng2022rlprompt} trains a policy network using reinforcement learning to generate prompts that maximize task accuracy. Prompt tuning methods such as prefix-tuning~\cite{li2021prefix} and soft prompts~\cite{lester2021power} optimize continuous embeddings rather than discrete text, but require access to model parameters. 
In contrast, our Bayesian approach treats the LLM as a black box and leverages a GP surrogate for sample-efficient exploration of the discrete prompt space.

ProTeGi is an algorithm that uses gradient-inspired edits and beam search to automatically optimize prompts. It generates candidate prompts from a seed by identifying errors on a small dataset, prompting an LLM to generate ``gradients" (i.e., critiques), and then using another LLM to apply these critiques to edit the prompt. It performed well on several binary classification tasks, including LIAR and ETHOS, which we use as a baseline. BO-LLM uses components from ProTeGi, particularly its expansion function, but replaces the heuristic search with a formal BO framework\cite{pryzant2023automaticpromptoptimizationgradient}.

\section{Methodology}
\begin{figure}[hbt!]
\centering
\includegraphics[width=\columnwidth]{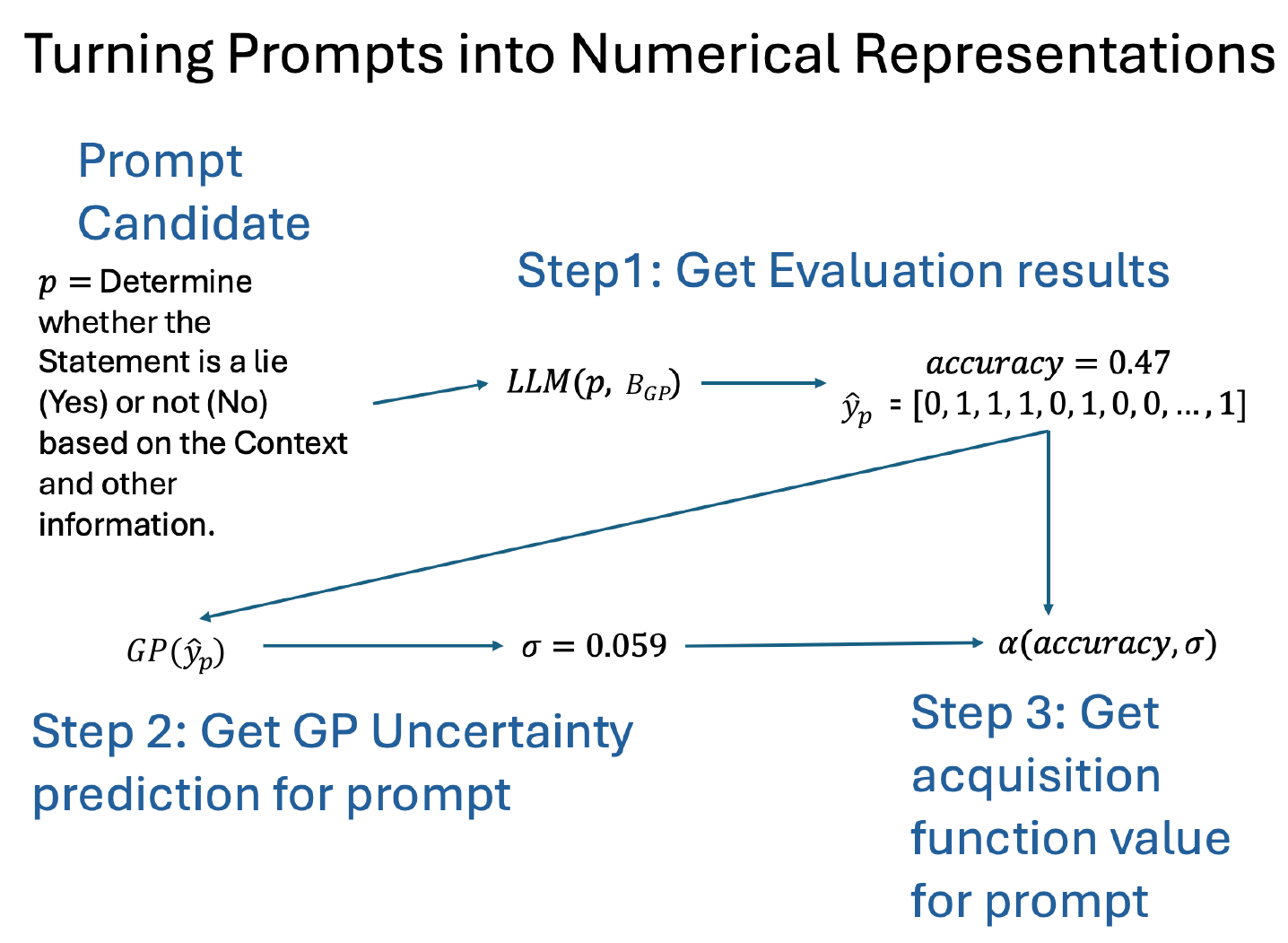}
\caption{How prompts are transformed from texts to numerical representation.}
\label{fig:num_representation}
\end{figure}

We propose BO-LLM, a framework that applies BO to prompt search. The objective is to identify prompts that maximize classification accuracy. Unlike heuristic methods, BO-LLM explicitly models the performance landscape of candidate prompts with a GP surrogate, enabling sample-efficient optimization. Figure~\ref{fig:num_representation} illustrates how prompts are transformed from text to numerical representations for optimization.

As illustrated in Fig.~\ref{fig:num_representation}, each textual prompt $p$ is turned into a numerical object in three complementary ways: 
(i) the prediction vector $\hat y_p$ on the fixed control minibatch $B_{GP}$, 
(ii) the surrogate’s posterior mean $\mu(p)$ and uncertainty $\sigma(p)$ computed from $\hat y_p$, and 
(iii) the acquisition score $\alpha(p)$ derived from $(\mu,\sigma)$. 
These representations jointly support sample-efficient selection during BO-LLM.

\subsection{Bayesian Optimization Formulation}

Let $\mathcal{P}$ denote the discrete prompt space. Given a labeled 
dataset $\mathcal{D}=\{(x_i,y_i)\}_{i=1}^N$, our objective is to 
maximize the test accuracy of an LLM queried with prompt $p\in\mathcal{P}$:
\begin{equation}
f(p)=\mathbb{E}_{(x,y)\sim \mathcal{D}}\big[\mathbbm{1}
\{\mathrm{LLM}_p(x)=y\}\big].
\end{equation}

Each evaluation on a subset of $\mathcal{D}$, i.e., $B_\alpha \subset \mathcal{D}$, yields a noisy observation 
$a(p) = f(p) + \varepsilon$, where $\varepsilon\sim\mathcal{N}
(0,\sigma_\varepsilon^2)$. We maintain a cache of observations 
$D=\{(p_i, a(p_i)\}_{i=1}^n$.

\paragraph{Data Representation} 
To build the surrogate model, each prompt $p$ is mapped to a 
prediction vector $\hat{y}_p \in \{0,1\}^{|B_{GP}|}$ on a fixed 
control minibatch $B_{GP} \subset \mathcal{D}$~\cite{brochu2010tutorialbayesianoptimizationexpensive}. 
This vector serves as the prompt's numerical representation. The 
similarity between two prompts $p_i$ and $p_j$ is measured via a 
Radial Basis Function (RBF) kernel:
\begin{equation}
k(p_i,p_j)=\exp\left(-\frac{\|\hat{y}_{p_i}-\hat{y}_{p_j}\|_2^2}{2R}\right)
\end{equation}
which captures the intuition that prompts with similar prediction 
patterns on $B_{GP}$ have correlated accuracies. The kernel 
hyperparameter $R$ was set empirically to 1.0 after preliminary runs 
indicated stable performance at this value. 

The mean of each prompt candidate's prediction accuracy regarding $B_{GP}$  is calculated as follows:
\begin{equation}~\label{eq:B_GP_mean}
m(p) = \frac{1}{|B_{GP}|}\sum_{i \in B_{GP}} \mathbbm{1}\{\hat{y}_i = y_i\}
\end{equation}
where $\hat{y}_i$ represents the predicted value for a given sentence and $y_i$ represents the ground truth label.

\paragraph{Surrogate Model}  
Given observations $D=\{(p_i,a(p_i))\}_{i=1}^n$, we update the GP 
prior over $f$ using Bayes' rule:
\begin{equation}
P(f\,|\,D)\ \propto\ P(D\,|\,f)\,P(f).
\end{equation}
Here $P(f)$ is the GP prior over functions and $P(D\,|\,f)$ is the 
Gaussian likelihood with observation noise. Conditioning the joint 
Gaussian yields the closed-form posterior used in our surrogate 
(see Appendix~\ref{app:gp-derivation}).

 We can further calculate the posterior mean for each prompt using:
\begin{equation}
\mu(p) = m(p) + \mathbf{k}_p^\top(\mathbf{K}+\sigma_\varepsilon^2
\mathbf{I})^{-1}(\mathbf{a} - \mathbf{m})
\end{equation}
where $\mathbf{k}_p = [k(p, p_1), \ldots, k(p, p_n)]^\top$ is the 
covariance vector between the new prompt $p$ and all observed prompts 
in $D$, $\mathbf{K}$ is the $n \times n$ kernel matrix with entries 
$K_{ij} = k(p_i, p_j)$, $\mathbf{a} = [a(p_1), \ldots, a(p_n)]^\top$ 
is the vector of observed accuracies of   the prompts in $D$ on  set $B_{\alpha}$, and $\mathbf{m} = [m(p_1), 
\ldots, m(p_n)]^\top$ is the vector of prediction accuracy of  the prompts in $D$ on  set $B_{GP}$ as given in~\eqref{eq:B_GP_mean}.

Given the observation history $D$, the GP posterior provides a 
predictive distribution for any new prompt. The posterior mean 
$\mu(p)$ and variance $\sigma^2(p)$ are calculated as follows:

\begin{equation}
\sigma^2(p) = k(p,p)-\mathbf{k}_p^\top(\mathbf{K}+\sigma_\varepsilon^2
\mathbf{I})^{-1}\mathbf{k}_p
\end{equation}

\paragraph{Acquisition Function}
After generating a set of new candidates, we use the GP posterior to 
estimate their mean accuracy $\mu(p)$ and uncertainty $\sigma(p)$. 
The Upper Confidence Bound acquisition function is the key 
component for balancing exploration and exploitation:
\begin{equation}
\alpha_{\text{UCB}}(p) = \mu(p) + \kappa\sigma(p)
\label{eq:ucb}
\end{equation}
where $\kappa > 0$ controls the exploration-exploitation trade-off. 
The first term $\mu(p)$ encourages \emph{exploitation} by favoring 
prompts with high predicted accuracy, while the second term 
$\kappa\sigma(p)$ encourages \emph{exploration} by favoring prompts 
with high uncertainty. We select the prompt $p$ that maximizes this 
score, evaluate its true accuracy on $B_\alpha$, and add the result 
to our observation data for the next round. UCB enjoys theoretical 
guarantees in the bandit setting and is particularly robust under 
evaluation noise~\cite{srinivas2010gaussian}.

In our implementation, $\kappa$ is annealed linearly from 2.0 to 0.5 
over the 10 optimization rounds, transitioning from exploration-heavy 
behavior in early rounds to exploitation-focused selection as the 
surrogate model becomes more confident.

\paragraph{Alternative Acquisition (EI)}
Besides UCB, Expected Improvement (EI) is a standard choice~\cite{snoek2012practical,shahriari2016taking}:
\[
\mathrm{EI}(p)=\big(\mu(p)-f^\star-\xi\big)\Phi(Z)+\sigma(p)\phi(Z),
\quad Z=\frac{\mu(p)-f^\star-\xi}{\sigma(p)}
\]
where $f^\star$ is the best observed accuracy, $\Phi/\phi$ are the 
standard normal CDF/PDF, and $\xi\!\ge\!0$ controls exploration. We 
report UCB results for its robustness under evaluation noise, but EI 
is drop-in compatible with our pipeline.

\paragraph{Surrogate Details and Model Parameter Learning}

We optimize kernel parameters $(R,\sigma^2)$ by maximizing the log marginal likelihood:
\[
\log p(\mathbf{a}|\theta)=-\tfrac12\mathbf{a}^\top K_\theta^{-1}\mathbf{a}-\tfrac12\log|K_\theta|-\tfrac{n}{2}\log 2\pi.
\]
Gradient:
\[
\frac{\partial \log p}{\partial \theta}
=\tfrac12\mathbf{a}^\top K_\theta^{-1}\tfrac{\partial K_\theta}{\partial \theta}K_\theta^{-1}\mathbf{a}
-\tfrac12\mathrm{tr}\!\left(K_\theta^{-1}\tfrac{\partial K_\theta}{\partial \theta}\right).
\]
A full derivation is in Appendix~\ref{app:gp-derivation}.

\subsection{Overall Algorithm}

The optimization loop of BO-LLM proceeds in rounds. 
At each round the surrogate GP is updated, new candidates 
are generated from current seeds, and the most promising 
candidate is selected via the acquisition function. The 
selected prompt is then evaluated and added to the cache. 
This process is summarized in Algorithm~\ref{alg:overall}.

\begin{algorithm}[ht]
\caption{BO-LLM: Overall Loop}\label{alg:overall}
\begin{algorithmic}[1]
\Require Seeds $S_0$, control batch $B_{GP} \subset \mathcal{D} $, evaluation batch $B_\alpha \subset \mathcal{D}$, rounds $T$
\For{$t=1$ to $T$}
    \State Fit GP on cache $D_{t-1}$
    \State $C_t\leftarrow \mathrm{Expand}(S_{t-1}, B_\alpha)$
    \State $p_t\leftarrow \arg\max_{p\in C_t}\mu_t(p)+\kappa_t\sigma_t(p)$
    \State Evaluate $a(p_t)$ on $B_\alpha$, update $D_t$
    \State Update seeds $S_t$
\EndFor
\end{algorithmic}
\end{algorithm}
The overall optimization loop is summarized in Algorithm 1. The process is iterative, where in each round $t$ (from 1 to $T$), we update our model based on the accumulated data. We maintain a cache, denoted as $D$, which stores all prompt-accuracy pairs evaluated so far. At the beginning of round $t$, we use the cache from the previous round, $D_{t-1}$, to fit the GP surrogate model. A new prompt $p_t$ is then selected and evaluated, and the cache is updated to $D_t$ by adding the new observation $(p_t, a(p_t))$.

\section{Implementation of BO-LLM}
In this section, we discuss the implementation of the BO-LLM algorithm.

\subsection{System Architecture}

\begin{figure}[hbt!]
\centering
\includegraphics[width=\columnwidth]{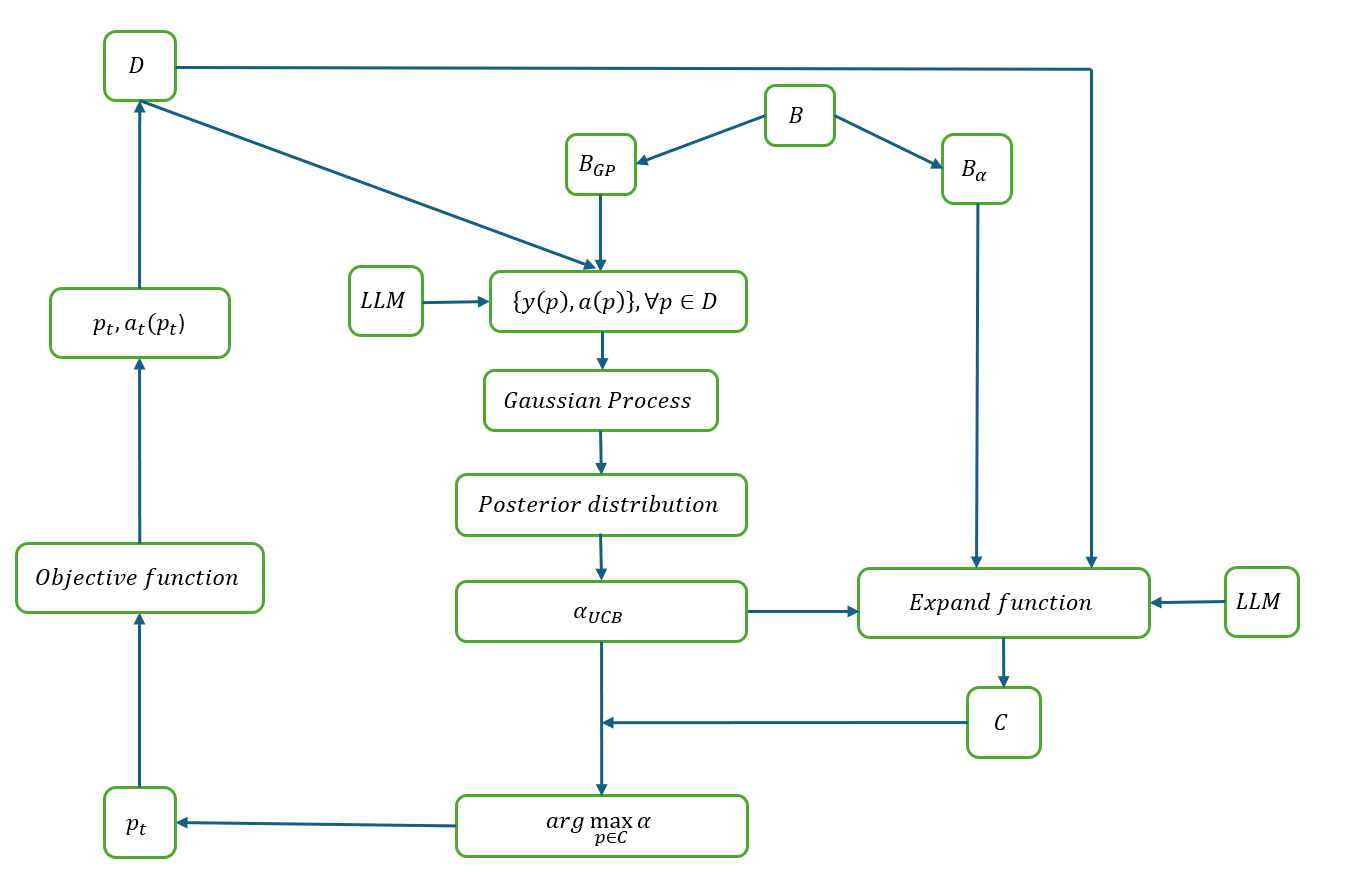}
\caption{Complete system architecture showing data flow between BO-LLM components.}
\label{fig:original_flowchart}
\end{figure}

Figure~\ref{fig:original_flowchart} presents the complete architecture of BO-LLM, illustrating the interaction between key components: the dataset $D$, the two evaluation minibatches ($B_{GP}$ and $B_\alpha$), the LLM-based expansion function, 
the GP surrogate, and the UCB acquisition mechanism. The bidirectional arrows indicate the iterative refinement process where selected prompts $p_t$ are evaluated on the objective function, and their observations are fed back to update the surrogate model.

\begin{algorithm}[h]
\caption{Expand function}\label{alg:expand}
\begin{algorithmic}[1]
    \Require $sp$: seed prompt; $B_\alpha \subset \mathcal{D}$: evaluation batch
    \State Evaluate prompt $p$ on $B_\alpha$ and collect errors: 
    $e=\{(x_i,y_i):(x_i,y_i) \in B_\alpha \wedge LLM_p(x_i)\neq y_i\}$
    \State Get gradient-inspired edits: $\{g_1,...,g_m\}=LLM_\nabla(sp,e)$
    \State Use edits to create new prompts: 
    $\{p'_{i1},...,p'_{iq}\}=LLM_\delta(p,g_i,e)$
    \State Get Monte Carlo successors for diversity: 
    $\{p''_{ij1},...,p''_{ijm}\}=LLM_{mc}(p'_{ij})$
    \State Remove seed prompt $sp$ to avoid re-testing
    \State \Return $\{p'_{11},...,p'_{mq}\} \cup \{p''_{111},...,p''_{mqp}\}$
\end{algorithmic}
\end{algorithm}

\subsection{Prompt Generation via Expansion}

The optimization loop generates new candidate prompts from high-performing seeds. The expansion process, outlined in Algorithm~\ref{alg:expand}, uses an LLM to generate edits inspired by classification errors on minibatch $B_\alpha$, combined with Monte Carlo rewrites for diversity.

Algorithm~\ref{alg:acquisition} integrates the expansion mechanism 
with the GP surrogate and UCB acquisition. Starting from the top-3 
seeds in the cache, it generates candidates via the expand function, 
evaluates each on the control minibatch $B_{GP}$ to obtain prediction vectors $\hat{y}_p$, and uses the GP posterior to compute acquisition scores. The candidate maximizing UCB is selected for evaluation on $B_\alpha$.

\begin{algorithm}[t]
\caption{Acquisition Optimization (UCB with GP over prediction vectors)}
\label{alg:acquisition}
\begin{algorithmic}[1]
\Require Posterior GP over prompts via prediction vectors on $B_{GP}$; 
         evaluation batch $B_\alpha \subset \mathcal{D}$; seed pool $\tilde D_t$
\State $C \gets \varnothing$ \Comment{candidate set}
\State Select $n{=}3$ best seeds by observed accuracy in $\tilde D_t$
\For{each seed $sp$}
  \State $e \gets \{(x,y)\!\in\!B_\alpha:\ \mathrm{LLM}_{sp}(x)\!\neq\!y\}$
  \State $\{g_i\}_{i=1}^4 \gets \mathrm{LLM}_\nabla(sp,e)$
  \For{each $g_i$}
    \State $p' \gets \mathrm{LLM}_\delta(sp,g_i,e)$
    \State $\{p''_j\}_{j=1}^2 \gets \mathrm{LLM}_{mc}(p')$
    \State $C \gets C \cup \{p', p''_1, p''_2\}$
  \EndFor
\EndFor
\For{$p \in C$}
  \State $\hat y_p \gets \mathrm{evaluate}(p, B_{GP})$
\EndFor
\State Compute $\mu(p),\sigma(p)$ from GP on $\{\hat y_p\}_{p\in C}$
\State \Return $p_t=\arg\max_{p\in C}\big[\mu(p)+\kappa\,\sigma(p)\big]$
\end{algorithmic}
\end{algorithm}

\FloatBarrier

\subsection{Flowchart}

\label{sec:flowchart}

\begin{figure}[hbt!]
\centering
\begin{tikzpicture}[node distance=6mm and 10mm, >=LaTeX,
  box/.style={rectangle, draw, rounded corners, align=center, 
              minimum width=0.75\columnwidth,
              inner sep=3pt, 
              minimum height=8mm}] 
\node[box] (seed) {Seed prompts};
\node[box, below=of seed] (expand) {Expand: edits + MC rewrites};
\node[box, below=of expand] (gp) {Fit GP $\Rightarrow$ $\mu,\sigma$};
\node[box, below=of gp] (ucb) {UCB select best candidate};
\node[box, below=of ucb] (eval) {Evaluate on $B_\alpha$};
\node[box, below=of eval] (update) {Update cache and seeds};

\draw[->, line width=0.8pt] (seed) -- (expand);
\draw[->, line width=0.8pt] (expand) -- (gp);
\draw[->, line width=0.8pt] (gp) -- (ucb);
\draw[->, line width=0.8pt] (ucb) -- (eval);
\draw[->, line width=0.8pt] (eval) -- (update);

\draw[->, line width=0.8pt] (update.east) -- ++(2.0,0) 
      node[pos=0.5, right] {} 
      |- (expand.east);
\end{tikzpicture}
\caption{BO-LLM optimization loop.}
\label{fig:bo_flow}
\end{figure}
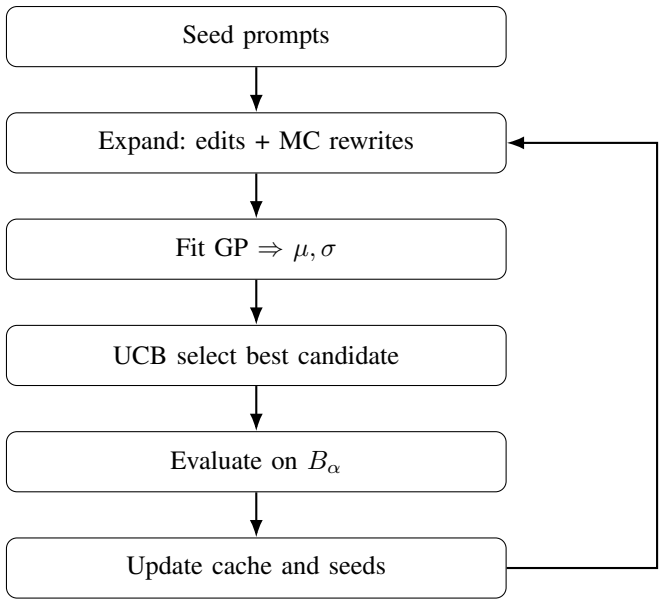

Figure~\ref{fig:bo_flow} summarizes the iterative optimization process of BO-LLM. 
The algorithm proceeds in a closed loop where each round consists of: (1) generating candidate prompts via the Expand function, (2) fitting a GP surrogate on cached observations, (3) selecting the highest-scoring candidate via UCB, (4) evaluating the selected prompt on minibatch $B_\alpha$, and (5) updating the cache and seed pool. The backward edge from "Update cache and seeds" to "Expand" represents the 
iterative refinement process that continues until convergence or a maximum number of rounds is reached.

\subsection{Additional Details}
During the expansion phase, each seed prompt generates candidates through a cascade of LLM operations with the following parameters:
\begin{itemize}[leftmargin=*,itemsep=2pt]
\item 4 gradient-inspired edits per seed (via $LLM_\nabla$)
\item 1 edit application step per gradient (via $LLM_\delta$)  
\item 2 Monte Carlo paraphrases per edited prompt (via $LLM_{mc}$)
\item Resulting in approximately 3 unique candidates per seed after deduplication
\end{itemize}

The prompts are evaluated using a Cached01Scorer that maintains a cache of previously evaluated prompt-accuracy pairs to avoid redundant API calls. The UCB hyperparameter $\kappa$ is annealed linearly from 2.0 to 0.5 over 10 rounds, transitioning from exploration to exploitation.

For the expansion function, we generate 4 $g$ gradients, with 1 gradient per error, 1 step per gradient, 2 Monte Carlo samples per stem, and 3 prompt generations per seed prompt. The evaluator is set to UCB and the scorer is set to the Cached01Scorer.

\subsection{Comparison with ProTeGi} 
ProTeGi proceeds in multiple candidates per round: gradient-inspired edits, beam search, and Monte-Carlo rewrites. It keeps top-$k$ as new seeds, improving stability but with $m$-fold higher API cost. BO-LLM differs by fitting a surrogate and selecting one candidate per round.

\paragraph{ProTeGi pipeline (gradient, edits, MC).}
ProTeGi~\cite{pryzant2023automaticpromptoptimizationgradient} operates in rounds starting from seed prompts $sp$ and an eval minibatch $B_\alpha$ with three LLM roles:
\begin{itemize}[leftmargin=*,itemsep=2pt,topsep=2pt]
\item $\mathbf{LLM}_\nabla$ (\emph{gradient-inspired edits}): consume $(sp,e)$ where $e=\{(x,y)\in B_\alpha:\ \mathrm{LLM}_{sp}(x)\neq y\}$ and propose textual “gradients” $g_1,\dots,g_m$;
\item $\mathbf{LLM}_\delta$ (\emph{edit applier}): apply $g_i$ to $sp$ to form candidates $p'_{i1},\dots,p'_{iq}$;
\item $\mathbf{LLM}_{mc}$ (\emph{Monte-Carlo rewrites}): paraphrase $p'_{ij}$ to explore local neighborhoods, producing $p''_{ij1},\dots,p''_{ijr}$.
\end{itemize}
Candidates are scored on a minibatch; a beam/top-$k$ policy keeps several high scorers as next-round seeds. This multi-candidate policy improves stability at the cost of higher API usage. BO-LLM instead fits a surrogate on prediction patterns and typically selects a single most-informative candidate via UCB.

\subsection{Practical Considerations}
\paragraph{Evaluation noise.}
Even with identical prompts, we observed 5--10\% variation in accuracy due to LLM nondeterminism. To mitigate this, BO-LLM applies a variance floor $\sigma_{\min}=0.02$ and allows repeated evaluation when the UCB margin is small.

\paragraph{Label-reversal pitfalls.}
Occasionally a candidate prompt implicitly reverses labels (e.g., mapping ``true'' to 0 and ``false'' to 1). Such prompts may achieve misleading minibatch accuracy and contaminate the seed pool. We detect these cases by checking confusion patterns against a reference prompt.

\paragraph{Stability vs.\ cost.}
Selecting only one candidate per round yields $\mathcal{O}(T)$ API calls, but is more prone to local maxima. Batch-UCB with $m=3$ improves stability but requires $\approx 3\times$ calls. 
This trade-off is analyzed further in our ablation studies.

\section{Experiments}

\subsection{Setup and Datasets}
We compare BO-LLM with ProTeGi on LIAR~\cite{wang2017liarliarpantsfire} and ETHOS~\cite{Mollas_2022}. The LIAR dataset contains 12.8k political statements labeled for truthfulness, which we map to a binary task (true vs. false). The ETHOS dataset contains over 8k social media comments labeled for hate speech (hate vs. non-hate).

Each algorithm is run for 10 rounds, starting with the same initial prompt for fair comparison. For BO-LLM, the control minibatch $B_{GP}$ has 75 items, and the evaluation minibatch $B_{\alpha}$ has 50 items. To mitigate potential biases from class imbalance, the control minibatch $B_{GP}$ was sampled to maintain the original dataset's label distribution. Accuracy is reported on held-out test sets, averaged over three runs.

\subsection{Evaluation Protocol}
We report test accuracy from three independent runs with different 
random seeds. Both BO-LLM and ProTeGi start from the same initial 
seed prompt for fair comparison. Each algorithm runs for $T = 10$ 
rounds with identical API budgets. For BO-LLM, the control minibatch 
$B_{GP}$ contains 75 items, and the evaluation minibatch $B_\alpha$ 
contains 50 items. 

For each run, we record the best accuracy achieved across all rounds. The reported results show the trajectory of the best prompt found up to each round, averaged across the three runs. We also report the standard deviation to capture the variance in LLM evaluations, which ranges from 5--10\% even for identical prompts due to model nondeterminism~\cite{atil2025nondeterminismdeterministicllmsettings}.
This protocol ensures a fair and statistically robust comparison between methods.

\subsection{Results}

\begin{figure}[hbt!]
\centering
\includegraphics[width=0.45\textwidth]{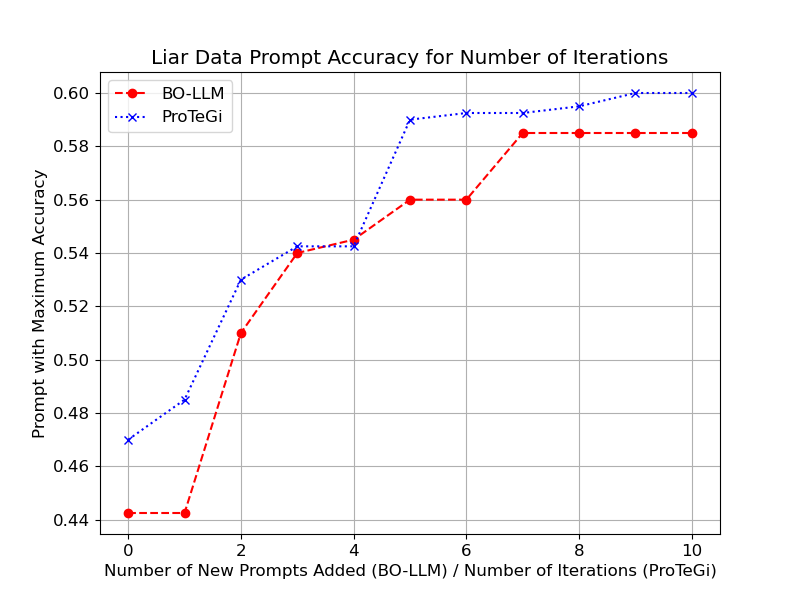}
\includegraphics[width=0.45\textwidth]{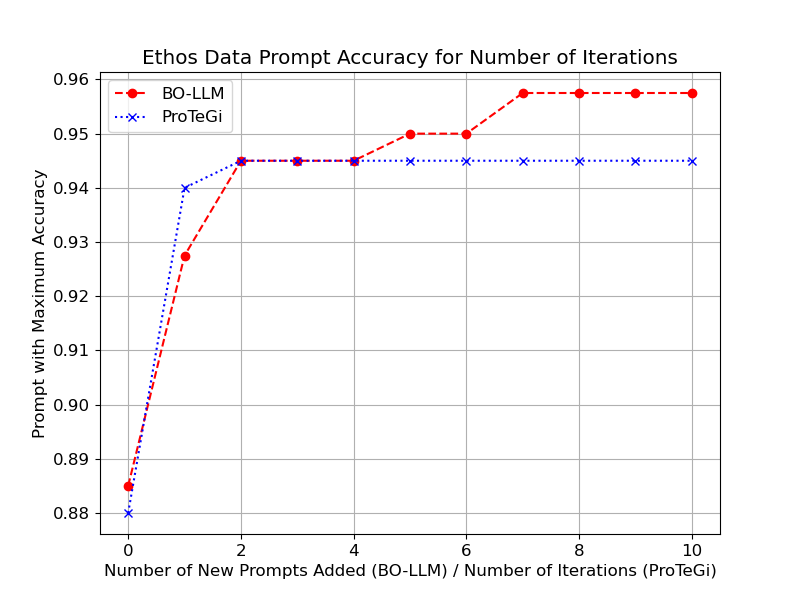}
\caption{Comparison of BO-LLM and ProTeGi on LIAR and ETHOS datasets over 10 rounds.}
\label{fig:liar_ethos}
\end{figure}

\paragraph{LIAR Dataset}
On the LIAR dataset, both algorithms show similar upward trends in accuracy over 10 rounds. ProTeGi reached a peak accuracy of \textbf{60.0\%} at round 9, while BO-LLM peaked at \textbf{58.5\%} at round 7. Compared to the initial prompt's performance, BO-LLM achieved a 32\% relative improvement versus 27.7\% for ProTeGi. The variance across rounds remains high for both, reflecting the ambiguous nature of political statements.

\paragraph{ETHOS Dataset}
On the ETHOS dataset, BO-LLM slightly outperformed ProTeGi, achieving a peak accuracy of \textbf{95.75\%} at round 7 compared to ProTeGi's \textbf{94.5\%} at round 2. The relative accuracy gains over the initial prompt were \textbf{8.2\%} for BO-LLM and \textbf{7.4\%} for ProTeGi. BO-LLM also appeared to converge more quickly, with accuracy stabilizing by round 5.

It must be noted that the initial prompt for the BO-LLM and ProTeGi algorithms are the same. The variability in LLM accuracy with the same prompt on the LIAR dataset can be seen on the 0th round as the two algorithms began with the same prompt yet the LLM obtained slightly different accuracies for it. This will therefore influence the relative accuracy increase both algorithms were able to achieve.

\subsection{Prompt Evolution Analysis}

Table~\ref{tab:prompt_evolution} shows how BO-LLM iteratively refines prompts on the LIAR dataset, incorporating task-relevant features that progressively improve accuracy.

\begin{table}[h]
\centering
\caption{Prompt evolution on LIAR dataset}
\label{tab:prompt_evolution}
\small
\begin{tabular}{@{}p{0.8cm}p{6cm}c@{}}
\toprule
\textbf{Iter.} & \textbf{Prompt} & \textbf{Acc.} \\
\midrule
0 & Determine whether the Statement is a lie (Yes) or not (No) based on the Context and other information. & 0.470 \\
\midrule
1 & Determine whether the Statement is a lie (Yes) or not (No) based on the Context, the individual's Job title, State, and Party affiliation. & 0.535 \\
\midrule
3 & Determine whether the Statement is a lie (Yes) or not (No) based on the source's job title, state, party affiliation, and context. Consider the credibility and bias of the source. & 0.540 \\
\midrule
5 & Decide if the Statement is false (Yes) or true (No) by considering the credibility and bias of the source, along with the content and factual accuracy. & 0.570 \\
\bottomrule
\end{tabular}
\end{table}

\paragraph{Key Findings}
The table shows a +21.3\% relative improvement from baseline (0.47) to iteration 5 (0.57). The algorithm progressively identifies relevant features: job title and party affiliation (Iter. 1), source credibility considerations (Iter. 3), and factual accuracy emphasis (Iter. 5).

\subsection{Analysis of Posterior Dynamics}
\begin{figure}[hbt!]
\centering
\includegraphics[width=0.9\columnwidth]{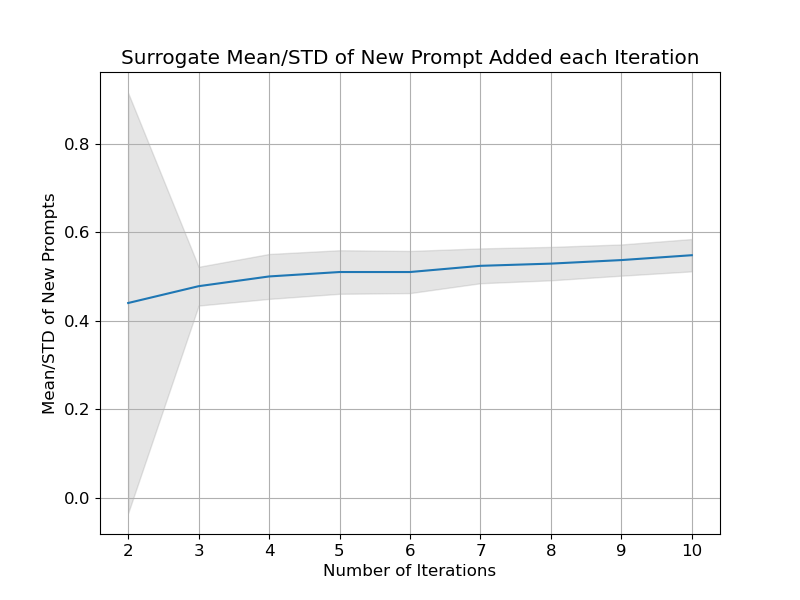}
\caption{Surrogate posterior mean and STD of prompts selected by UCB (LIAR dataset).}
\label{fig:surrogate_mean}
\end{figure}

To better understand the optimization dynamics of BO-LLM, we analyze the surrogate model's posterior mean and variance. Figure~\ref{fig:surrogate_mean} illustrates this process on the LIAR dataset.

We observe that the posterior variance decreases rapidly during the first three rounds and then tapers off, reflecting the transition from exploration to exploitation. This indicates the surrogate model quickly gains confidence. Concurrently, the posterior mean continues to rise, showing that the acquisition strategy successfully identifies increasingly better prompts. However, this also reveals a potential failure mode: if the variance shrinks too quickly around a misleading candidate, the optimization can converge to a suboptimal plateau.

\subsection{Multi-Turn Extension}
We instantiate BO-LLM on a clarification QA pattern. 
Given an ambiguous user query $q$ (e.g., ``Tell me about Apple''), a good system prompt should first ask a clarifying question before answering.
We define a binary score $s(p,q)\!\in\!\{0,1\}$ that equals $1$ if the model’s \emph{first} turn asks for clarification (detected by a heuristic/regex or a lightweight classifier), and $0$ otherwise.
For a fixed set of ambiguous queries $\mathcal{Q}$, the objective is
\[
f(p)=\frac{1}{|\mathcal{Q}|}\sum_{q\in \mathcal{Q}} s(p,q).
\]
BO-LLM proceeds unchanged: candidates from \emph{Expand} are run on a control subset $B_{GP}\!\subset\!\mathcal{Q}$ to obtain prediction vectors $\hat y_p$; the GP posterior $(\mu,\sigma)$ feeds UCB (or EI) to select the next candidate; evaluation uses a disjoint subset $B_\alpha\!\subset\!\mathcal{Q}$.
A minimal seed template is:
\begin{quote}\small
``You will first \textbf{ask exactly one clarifying question} if the user's intent is ambiguous; only after receiving the answer, you will respond to the original request.''
\end{quote}
This demonstrates how BO-LLM extends beyond single-turn classification to instruction-level, procedural behaviors.

\section{Discussion and Conclusion}

\subsection{Discussion}
Our experiments show that BO-LLM achieves competitive accuracy, but its performance is subject to several factors. A central theme is the trade-off between stability and efficiency.

\paragraph{Failure Modes} A notable failure mode is the generation of "reversed-label" prompts, which unintentionally flip the intended labeling scheme (e.g., answering "true" with a "0" and "false" with a "1"). An example prompt that caused this was \textit{"Determine if this statement is true or false depending on the context included."}. While usually scoring poorly, such a prompt can occasionally achieve high minibatch accuracy by chance. If selected early, it can poison the seed pool and lead to suboptimal local maxima. Because BO-LLM evaluates only a single candidate per iteration, it may be more susceptible to this than multi-candidate strategies like ProTeGi.
This phenomenon was particularly true for the LIAR dataset. During the expansion portion of the algorithm, it was not uncommon for the LLM to generate a prompt that reversed the answering of each statement. For example, if the prompt was required to assign 1 for false statements and 0 for true statements, it would generate a prompt that would answer the opposite. This was normally not a problem as the prompt would score poorly upon evaluation. Occasionally, however, the prompt would gain a relatively high accuracy, and as a result would be used as a seed prompt. This was more likely if such a prompt was selected earlier on. The BO-LLM algorithm may be more susceptible to this when compared to ProTeGi based on only one prompt being selected each round for BO-LLM while ProTeGi selects 4. This phenomenon was not observed with the ETHOS dataset.

\paragraph{Illustrative Example of Reversed-Label Prompts.} 
\begin{verbatim}
Correct: "Classify the following statement 
as True or False:"
Reversed-label: "Respond 'True' if the
statement is false, 
and 'False' if the statement is true:"
\end{verbatim}
The latter achieves misleading minibatch accuracy but poisons the surrogate.

\paragraph{Evaluation Variance} We observed high variability in LLM evaluations, even for identical prompts. For the initial prompt on the LIAR dataset, the starting accuracy sometimes differed by as much as 10\% between runs. This nondeterminism, also noted in other work~\cite{atil2025nondeterminismdeterministicllmsettings, zheng2023judging}, introduces noise that can bias the surrogate model and mislead the acquisition function.
The variation in evaluation with the same prompt was noticed as well. Both algorithms began with the same starter prompt for both datasets, yet the LLM used to evaluate the initial prompt would assign accuracies with as much as a 10\% difference in some cases. This instability was present within the performance of both classification tasks.

\paragraph{Trade-offs in Stability and Efficiency} To mitigate getting stuck in local maxima, one could evaluate multiple candidates per round instead of just one. While this increases stability, it multiplies the number of expensive API calls, sacrificing sample efficiency]. There is an inherent trade-off between the robustness gained from more evaluations and the cost-efficiency of a single-candidate approach, which is worth investigating in future studies.

\paragraph{API Cost Considerations.} 
Evaluating $m$ candidates per round yields $\mathcal{O}(mT)$ API calls vs. $\mathcal{O}(T)$ for single-candidate BO-LLM. 
For academic datasets, $m=3$ or $5$ is affordable and stabilizes learning. 
In production with costly APIs, sample-efficient single-candidate BO-LLM is preferable. 
The balance depends on application needs: robustness-critical vs. budget-constrained settings.

\paragraph{Complexity and stability.}
Full GP updates cost $\mathcal{O}(n^3)$ with $n$ observations; rank-one Cholesky updates amortize to $\mathcal{O}(n^2)$ per round.
After factorization, scoring $|C_t|$ candidates is $\mathcal{O}(|C_t|\,n)$.
Stability can be improved with small batch-UCB (evaluate top-$m$ per round), repeated evaluations with majority voting, a floor on predictive variance to avoid premature collapse, and periodic random restarts of seed prompts.

\subsection{Limitations and Future Work}
Our study is limited to two binary classification datasets. However, the framework naturally extends to multi-class problems. This can be achieved by modifying the prediction vector $\hat{y}_p$ from a binary value to a one-hot encoding of the predicted class, with corresponding adjustments to the distance metric in the kernel.

The GP surrogate's computational complexity ($O(n^3)$ with $n$ observations) also poses challenges for scaling. Exact GP updates scale cubically with the number of observations, making large-scale runs computationally challenging. Evaluation variance across identical prompts introduces noise into the surrogate and acquisition steps. In addition, relying on a single candidate per round increases vulnerability to local maxima compared to multi-candidate approaches.

Future work should extend BO-LLM to more diverse tasks, such as summarization and dialogue. We plan to investigate stabilization strategies like batch-UCB selection and variance reduction techniques. Exploring alternative acquisition functions and more scalable surrogates are also promising directions. These limitations highlight future directions such as batch selection, variance reduction, and more scalable surrogate models.

\subsection{Conclusion}
This paper demonstrates that BO-LLM, a framework applying BO to prompt search, achieves performance comparable to the heuristic-based ProTeGi. The primary advantage of the BO framework is its potential for sample efficiency, guided by a probabilistic surrogate model. However, it is susceptible to local maxima, particularly on noisy datasets, and is sensitive to the high variance of LLM evaluations. Despite these drawbacks, this work illustrates that BO is a viable and principled framework for automated prompt engineering, offering a distinct, model-based alternative to heuristic search methods.
BO-LLM, however, was susceptible to more frequent problems. When testing on the LIAR dataset, BO-LLM would occasionally find itself in a local maximum in which the prompt would not be specific enough to answer in the correct way, yet occur in an earlier round making it a seed prompt for later rounds. This may be caused by the fact that a single prompt is selected for each round with BO-LLM.

\appendix


\subsection{Dataset Descriptions}

\textbf{LIAR Dataset.} The LIAR dataset consists of 12,836 short statements from various political contexts~\cite{wang2017liarliarpantsfire}. We collapse the original six truthfulness categories into a binary label (true vs. false). An example is:
\begin{verbatim}
{"label": 1, "text": "Statement: Says the 
Annies List political group supports third-
trimester abortions on demand.\nJob title: 
State representative\nState: Texas\nParty: 
republican\nContext: a mailer"}
\end{verbatim}

\textbf{ETHOS Dataset.} The ETHOS dataset contains over 8,000 English social media comments labeled for hate speech detection~\cite{Mollas_2022}. We use the binary (hate vs. non-hate) setting. An example is:
\begin{verbatim}
Gosh you guys are getting less funny with 
each episode.;0.0
\end{verbatim}

\subsection{Hyperparameters and Sketch of GP Posterior Derivation}\label{app:gp-derivation}
We run each algorithm for 10 rounds, using the same initial seed prompts for both methods. For BO-LLM, the control minibatch $B_{GP}$ size is 75 and the evaluation minibatch $B_{\alpha}$ size is 50. During expansion, we generate 4 gradient-inspired edits, 1 edit step per gradient, and 2 Monte Carlo samples, producing 3 new candidates per seed prompt. The UCB hyperparameter $\kappa$ is annealed from 2.0 to 0.5. All experiments are repeated with 3 random seeds, and we report the average of the highest-performing prompt's accuracy at each round.

Let $\mathbf{a}=[a(p_1),...,a(p_n)]^\top$, $f_\star=f(p_\star)$, $m_* = m(p_*)$, and  $\mathbf{k}_* = [k(p_*, p_1), \ldots, k(p_*, p_n)]^\top$ .  
\[
\begin{bmatrix}\mathbf{a}\\ f_\star\end{bmatrix}\!\sim\!\mathcal{N}\!\left(
\begin{bmatrix} \mathbf{m} \\ m_{*} \end{bmatrix} ,\begin{bmatrix}K+\sigma^2I & \mathbf{k}_\star\\ \mathbf{k}_\star^\top & k(p_\star,p_\star)\end{bmatrix}\right).
\]
Conditioning gives the posterior mean and variance in the main text.

\bibliographystyle{IEEEtran}

\bibliography{mybib}

\end{document}